\documentclass[11pt,a4paper]{article}
\usepackage[hyperref]{acl2019}
\usepackage{times}
\usepackage{latexsym}
\usepackage{graphicx}
\usepackage{multirow}
\usepackage{amsfonts}
\usepackage{subfigure}
\usepackage{bbm}
\usepackage{amsmath}
\usepackage{booktabs}
\usepackage{url}
\usepackage{enumitem}
\usepackage{url}

\aclfinalcopy 
\def\aclpaperid{230} 

\title{Incremental Learning from Scratch for Task-Oriented Dialogue Systems}

\author{Weikang Wang$^{1,2}$, 
	Jiajun Zhang$^{1,2}$, 
	Qian Li$^3$,\\
	\textbf{Mei-Yuh Hwang$^3$,
	Chengqing Zong$^{1,2,4}$ \and 
	Zhifei Li$^3$} \\
	$^1$ National Laboratory of Pattern Recognition, Institute of Automation, CAS, Beijing, China \\ 
	$^2$ University of Chinese Academy of Sciences, Beijing, China \\
	$^3$ Mobvoi, Beijing, China \\
	$^4$ CAS Center for Excellence in Brain Science and Intelligence Technology, Beijing, China  \\
	{\tt \{weikang.wang, jjzhang, cqzong\}@nlpr.ia.ac.cn} \\ 
	{\tt \{qli, mhwang, zfli\}@mobvoi.com}}

\date{}

\begin{document}
\maketitle
\begin{abstract}
	Clarifying user needs is essential for existing task-oriented dialogue systems. However, in real-world applications, developers can never guarantee that all possible user demands are taken into account in the design phase. Consequently, existing systems will break down when encountering unconsidered user needs. To address this problem, we propose a novel incremental learning framework to design task-oriented dialogue systems, or for short \textbf{I}ncremental \textbf{D}ialogue \textbf{S}ystem~(\textbf{IDS}), without pre-defining the exhaustive list of user needs. Specifically, we introduce an \emph{uncertainty estimation module} to evaluate the confidence of giving correct responses. If there is high confidence, IDS will provide responses to users. Otherwise, humans will be involved in the dialogue process, and IDS can learn from human intervention through an \emph{online learning module}. To evaluate our method, we propose a new dataset which simulates unanticipated user needs in the deployment stage. Experiments show that IDS is robust to unconsidered user actions, and can update itself online by smartly selecting only the most effective training data, and hence attains better performance with less annotation cost.\footnote{\url{https://github.com/Leechikara/Incremental-Dialogue-System}}
\end{abstract}

\section{Introduction}
\begin{figure}[!t]
	\centering
	\setlength{\abovecaptionskip}{0.1cm}
	\setlength{\belowcaptionskip}{-0.1cm}
	\includegraphics[width=0.46\textwidth]{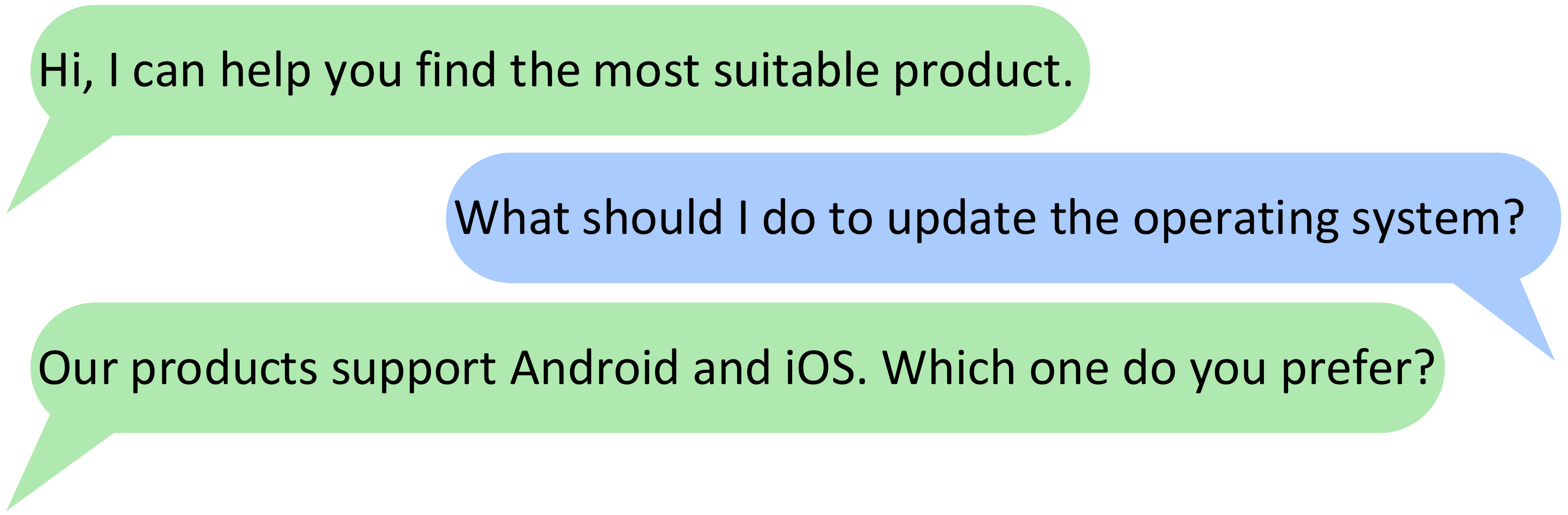}
	\caption{An example of task-oriented dialogue system. The system is designed to guide users to find a suitable product. Thus, when encountering unconsidered user needs such as "how to update the operating system", the system will give unreasonable responses.}
	\label{fig:fig1}
\end{figure}

Data-driven \emph{task-oriented dialogue systems} have been a focal point in both academic and industry research recently. Generally, the first step of building a dialogue system is to clarify what users are allowed to do. Then developers can collect data to train dialogue models to support the defined capabilities. Such systems work well if all possible combinations of user inputs and conditions are considered in the training stage~\cite{paek2008automating,wang2018teacher}. However, as shown in Fig.~\ref{fig:fig1}, if users have unanticipated needs, the system will give unreasonable responses.

This phenomenon is mainly caused by a biased understanding of real users. In fact, before system deployment, we do not know what the customers will request of the system. In general, this problem can be alleviated by more detailed user studies. But we can never guarantee that all user needs are considered in the system design. Besides, the user inputs are often diverse due to the complexity of natural language. Thus, it is impossible to collect enough training samples to cover all variants. Consequently, the system trained with biased data will not respond to user queries correctly in some cases. And these errors can only be discovered after the incident.

Since the real user behaviors are elusive, it is obviously a better option to make no assumptions about user needs than defining them in advance. To that end, we propose the novel \textbf{I}ncremental \textbf{D}ialogue \textbf{S}ystem~(\textbf{IDS}). Different from the existing training-deployment convention, IDS does not make any assumptions about the user needs and how they express intentions. In this paradigm, all reasonable queries related to the current task are legal, and the system can learn to deal with user queries online.

Specifically, after the user sends a query to our system, we use an \emph{uncertainty estimation module} to evaluate the confidence that the dialogue model can respond correctly. If there is high confidence, IDS will give its response to the user. Otherwise, human will intervene and provide a reasonable answer. When humans are involved, they can select a response from the current response candidates or give a new response to the user. If a new answer is provided, we add it to the system response candidates. Then, the generated context-response pair from humans will be fed into the dialogue model to update the parameters by an \emph{online learning module}. Through continuous interactions with users after deployment, the system will become more and more knowledgeable, and human intervention will become less and less needed.

To evaluate our method, we build a new dataset consisting of five sub-datasets~(named SubD1, SubD2, SubD3, SubD4 and SubD5) within the context of customer services. Following the existing work~\cite{bordes2016learning}, our dataset is generated by complicated and elaborated rules. SubD1 supports the most limited dialogue scenarios. Then each later sub-dataset covers more scenarios than its previous one. To simulate the unanticipated user needs, we train the dialogue models on simpler datasets and test them on the harder ones. Extensive experiments show that IDS is robust to the unconsidered user actions and can learn dialogue knowledge online from scratch. Besides, compared with existing methods, our approach significantly reduces annotation cost.

In summary, our main contributions are three-fold: (1) To the best of our knowledge, this is the first work to study the incremental learning framework for task-oriented dialogue systems. In this paradigm, developers do not need to define user needs in advance and avoid collecting biased training data laboriously. (2) To achieve this goal, we introduce IDS which is robust to new user actions and can extend itself online to accommodate new user needs. (3) We propose a new benchmark dataset to study the inconsistency of training and testing in task-oriented dialogue systems.

\section{Background and Problem Definition}
Existing work on data-driven task-oriented dialogue systems includes generation based methods~\cite{wen2016network,Eric2017Key} and retrieval based methods~\cite{bordes2016learning,williams2017hybrid,li2017end}. In this paper, we focus on the retrieval based methods, because they always return fluent responses.

In a typical retrieval based system, a user gives an utterance $x_t$ to the system at the $t\text{-}th$ turn. Let $(x_{t,1},...,x_{t,N})$ denote the tokens of $x_t$. Then, the system chooses an answer $y_t=(y_{t,1},...,y_{t,M})$ from the candidate response set $R$ based on the conditional distribution $p(y_t|C_t)$, where $C_t=(x_1,y_1,...,x_{t-1},y_{t-1},x_t)$ is the dialogue context consisting of all user utterances and responses up to the current turn.

By convention, the dialogue system is designed to handle predefined user needs. And the users are expected to interact with the system based on a limited number of dialogue actions. However, predefining all user demands is impractical and unexpected queries may be given to the system after the system is deployed. In this work, we mainly focus on handling this problem.

\section{Incremental Dialogue System}
\begin{figure}[!h]
	\centering
	\setlength{\abovecaptionskip}{0.1cm}
	\setlength{\belowcaptionskip}{-0.2cm}
	\includegraphics[width=0.48\textwidth]{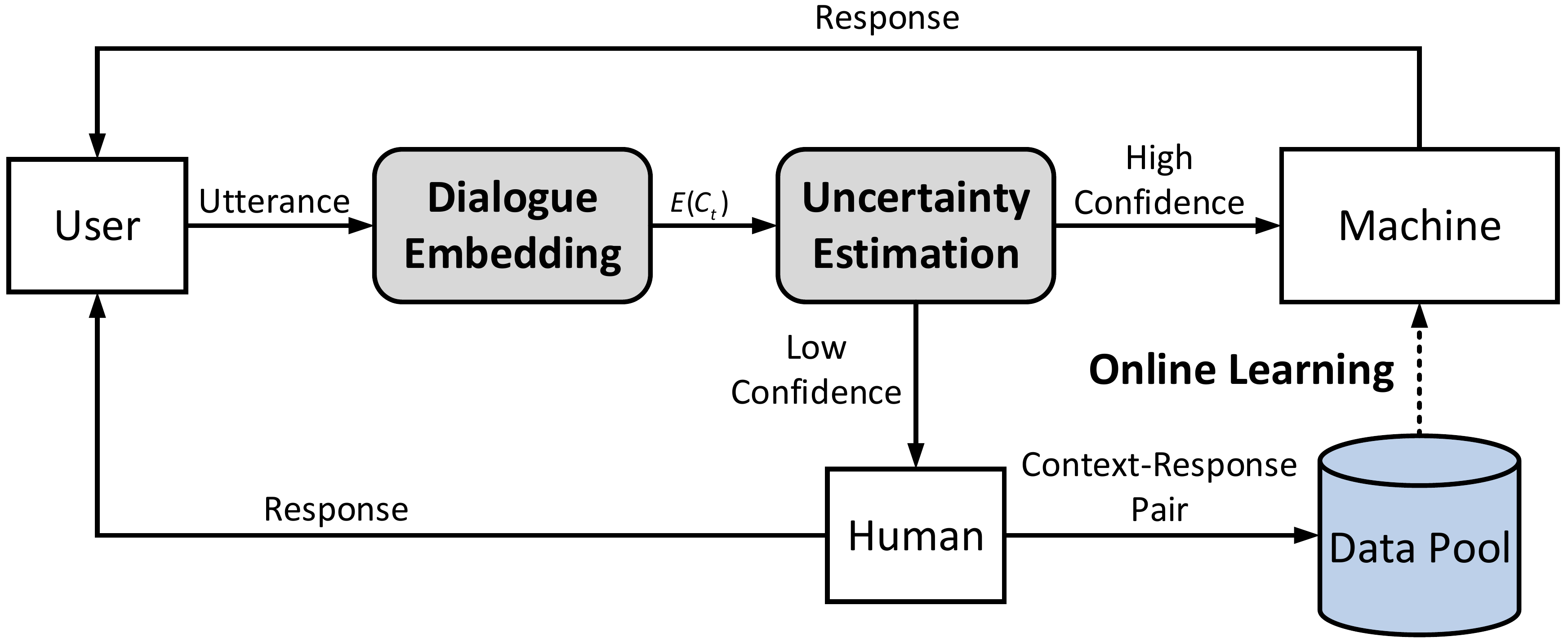}
	\caption{An overview of the proposed IDS.}
	\label{fig:fig2}
\end{figure}

As shown in Fig.~\ref{fig:fig2}, IDS consists of three main components: \emph{dialogue embedding module}, \emph{uncertainty estimation module} and \emph{online learning module}.

In the context of customer services, when the user sends an utterance to the system, the \emph{dialogue embedding module} will encode the current context into a vector. Then, the \emph{uncertainty estimation module} will evaluate the confidence of giving a correct response. If there is high confidence, IDS will give its response to the user. Otherwise, the hired customer service staffs will be involved in the dialogue process and provide a reasonable answer, which gives us a new ground truth context-response pair. Based on the newly added context-response pairs, the system will be updated via the \emph{online learning module}.

\subsection{Dialogue Embedding}
Given dialogue context $C_t$ in the $t\text{-}th$ turn, we first embed each utterance in $C_t$ using a Gated Recurrent Unit (GRU)~\cite{chung2014empirical} based bidirectional recurrent neural networks~(bi-RNNs). The bi-RNNs transform each utterance\footnote{We use $x$ to represent each user utterance and $y$ for each response for simplicity. All utterances use the same encoder.} $x=(w_1,w_2,...,w_N)$ in $C_t$ into hidden representation $H=({h}_1,{h}_2,...,{h}_N)$ as follows:
\begin{equation}
	\small
	\setlength{\abovedisplayskip}{0.35cm}
	\setlength{\belowdisplayskip}{0.35cm}
	\begin{split}
		\overrightarrow{h}_{n} &= \text{GRU}(\overrightarrow{h}_{n-1}, \phi^\text{emb}(w_{n})) \\
		\overleftarrow{h}_{n}  &= \text{GRU}(\overleftarrow{h}_{n+1}, \phi^\text{emb}(w_{n})) \\
		{h}_n                  &= \overrightarrow{h}_{n}\oplus\overleftarrow{h}_{n}
	\end{split}
\end{equation}
where $\phi^\text{emb}(w_{n})$ is the embedding of word $w_{n}$.

To better encode a sentence, we use the self-attention layer~\cite{lin2017structured} to capture information from critical words. For each element ${h}_n$ in bi-RNNs outputs, we compute a scalar self-attention score as follows:
\begin{equation}
	\small
	\setlength{\abovedisplayskip}{0.35cm} 
	\setlength{\belowdisplayskip}{0.35cm}
	\begin{split}
		a_n &= \text{MLP}({h}_n)\\
		p_n &= \text{softmax}(a_n)
	\end{split}
\end{equation}

The final utterance representation $E(x)$ is the weighted sum of bi-RNNs outputs:
\begin{equation}
	\small
	\setlength{\abovedisplayskip}{0.35cm} 
	\setlength{\belowdisplayskip}{0.35cm}
	E(x)= \sum_{n}p_n{h}_n
\end{equation}

After getting the encoding of each sentence in $C_t$, we input these sentence embeddings to another  GRU-based RNNs to obtain the context embedding $E(C_t)$ as follows:
\begin{equation}
	\label{Eq4}
	\small
	\setlength{\abovedisplayskip}{0.35cm} 
	\setlength{\belowdisplayskip}{0.35cm}
	E(C_t) = \text{GRU}(E(x_1),E(y_1),...,E(y_{t-1}),E(x_t))
\end{equation}

\subsection{Uncertainty Estimation}
In the existing work~\cite{williams2017hybrid,bordes2016learning,li2017end}, after getting the context representation, the dialogue system will give a response $y_t$ to the user based on $p(y_t|C_t)$. However, the dialogue system may give unreasonable responses if unexpected queries happen. Thus, we introduce the uncertainty estimation module to avoid such risks.

To estimate the uncertainty, we decompose the response selection process as follows:
\begin{equation}
	\label{Eq5}
	\small
	\setlength{\abovedisplayskip}{0.35cm} 
	\setlength{\belowdisplayskip}{0.35cm}
	p(y_t|C_t) = \int p(y_t|z,C_t)p(z|C_t)dz
\end{equation}

\begin{figure}[!t]
	\centering
	\setlength{\abovecaptionskip}{0.1cm}
	\setlength{\belowcaptionskip}{-0.2cm}
	\includegraphics[width=0.4\textwidth]{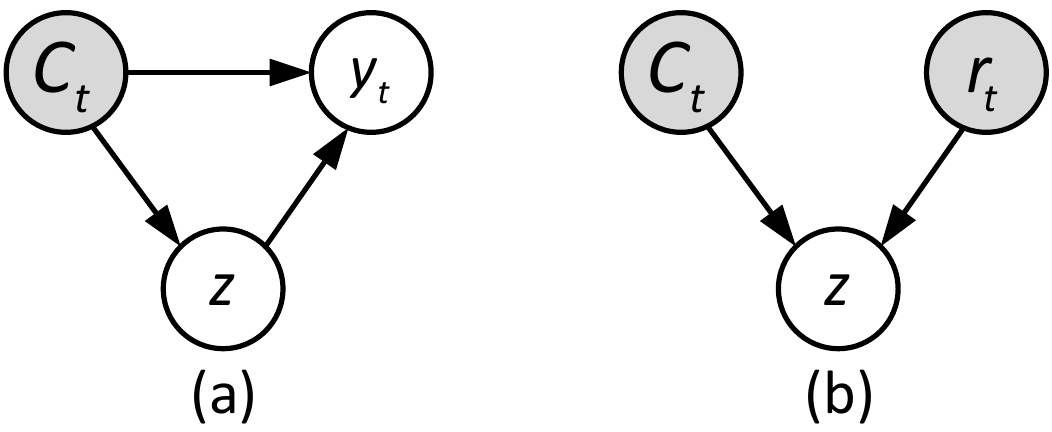}
	\caption{Graphical models of (a) response selection, and (b) online learning. The gray and white nodes represent the observed and latent variables respectively.}
	\label{fig:fig3}
\end{figure}

As shown in Fig.~\ref{fig:fig3}(a), from the viewpoint of probabilistic graphical models~\cite{koller2009probabilistic}, the latent variable $z$ can be seen as an explanation of the dialogue process. In an abstract sense, given $C_t$, there is an infinite number of paths $z$ from $C_t$ to $y_t$. And $p(y_t|C_t)$ is an expectation of $p(y_t|z,C_t)$ over all possible paths. If the system has not seen enough instances similar to $C_t$ before, the encoding of $C_t$ will be located in an unexplored area of the dialogue embedding space. Thus, the entropy of prior $p(z|C_t)$ will be large. If we sample latent variable $z$ based on $p(z|C_t)$ multiple times and calculate $p(y_t|z,C_t)$, we can find $p(y_t|z,C_t)$ has a large variance under different sampled latent variables $z$.

Based on such intuitive analysis, we design the uncertainty measurement for IDS. Specifically, we assume that the latent variable $z$ obeys a multivariate diagonal Gaussian distribution. Following the reparametrization trick~\cite{Kingma2014Auto}, we sample $\epsilon\sim\mathcal{N}(0,\mathbf{I})$ and reparameterize $z=\mu+\sigma\cdot\epsilon$. The mean and variance of the prior $p(z|C_t)$ can be calculated as:
\begin{equation}
	\small
	\setlength{\abovedisplayskip}{0.35cm} 
	\setlength{\belowdisplayskip}{0.35cm}
	\left[\begin{array}{c} 
		\mu \\
		log(\sigma^2) \\
	\end{array}\right] = \text{MLP}(E(C_t))
\end{equation}

After sampling a latent variable $z$ from the prior $p(z|C_t)$, we calculate the response probability for each element in the current candidate response set $R$. In IDS, $R$ will be extended dynamically. Thus, we address the response selecting process with the ranking approach. For each response candidate, we calculate the scoring as follows:
\begin{equation}
	\small
	\setlength{\abovedisplayskip}{0.35cm} 
	\setlength{\belowdisplayskip}{0.35cm}
	\begin{split}
		\rho(y_t|z,C_t) &= (E(C_t)\oplus z)^T \mathbf{W} E(y_t) \\
		p(y_t|z,C_t) &= \text{softmax}(\rho(y_t|z,C_t))
	\end{split}
\end{equation}
where $E(y_t)$ is the encoding of $y_t\in R$, and $\mathbf{W}$ is the weight matrices.

\begin{figure}[!t]
	\centering
	\setlength{\abovecaptionskip}{0.1cm}
	\setlength{\belowcaptionskip}{-0.2cm}
	\includegraphics[width=0.45\textwidth]{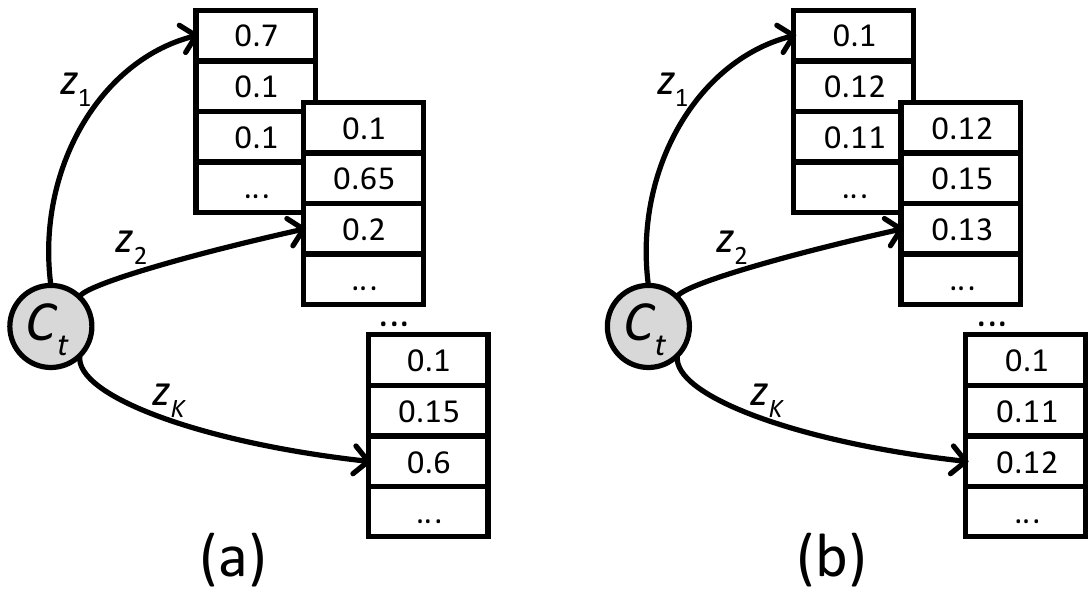}
	\caption{A toy example to show the uncertainty estimation criterions. (a) means a large variance in the response probability under different sampled latent variables. (b) means close weights to all response candidates in the early stage of online learning.}
	\label{fig:fig4}
\end{figure}

To estimate the variance of $p(y_t|z,C_t)$ under different sampled latent variables, we repeat the above process $K$ times. Assume that the probability distribution over the candidate response set in the $k\text{-}th$ repetition is $P_k$ and the average response probability distribution of $K$ sampling is $P_{avg}$. We use the Jensen-Shannon divergence ($\text{JSD}$) to measure the distance between $P_k$ and $P_{avg}$ as follows:
\begin{equation}
	\small
	\setlength{\abovedisplayskip}{0.35cm} 
	\setlength{\belowdisplayskip}{0.35cm}
	\text{JSD}(P_k||P_{avg}) = \frac{1}{2}(\text{KL}(P_k||P_{avg}) + \text{KL}(P_{avg}||P_k))
\end{equation}
where $\text{KL}(P||Q)$ is the Kullback\text{-}Leibler divergence between two probability distributions. Then, we get the average $\text{JSD}$ as follows:
\begin{equation}
	\small
	\setlength{\abovedisplayskip}{0.35cm} 
	\setlength{\belowdisplayskip}{0.35cm}
	\text{JSD}_{avg}=\frac{1}{K}\sum_{k=1}^K\text{JSD}(P_k||P_{avg})
\end{equation}
Because the average $\text{JSD}$ can be used to measure the degree of divergence of $\{P_1,P_2,...,P_K\}$, as shown in Fig.~\ref{fig:fig4}(a), the system will refuse to respond if $\text{JSD}_{avg}$ is higher than a threshold $\tau_{1}$.

However, the dialogue model tends to give close weights to all response candidates in the early stage of training, as shown in Fig.~\ref{fig:fig4}(b). It results in a small average $\text{JSD}$ but the system should refuse to respond. Thus, we add an additional criterion for the uncertainty measurement. Specifically, if the maximum probability in $P_{avg}$ is lower than a threshold $\tau_{2}$, the system will refuse to respond.

\subsection{Online Learning}
If the confidence is high enough, IDS will give the response with the maximum score in $P_{avg}$ to the user. Otherwise, the hired customer service staffs will be asked to select an appropriate response from the top T response candidates of $P_{avg}$ or propose a new response if there is no appropriate candidate. If a new response is proposed, it will be added to $R$. We denote the human response as $r_t$. Then, we can observe a new context-response pair $d_t=(C_t,r_t)$ and add it to the training data pool.

The optimization objective is to maximize the likelihood of the newly added data $d_t$. However, as shown in Eq.~\ref{Eq5}, calculating the likelihood requires an intractable marginalization over the latent variable $z$. Fortunately, we can obtain its lower bound~\cite{hoffman2013stochastic,miao2016neural,sohn2015learning} as follows:
\begin{equation}
	\label{Eq8}
	\small
	\setlength{\abovedisplayskip}{0.35cm} 
	\setlength{\belowdisplayskip}{0.35cm}
	\begin{split}
		\mathcal{L} &= \mathbb{E}_{q(z|d_t)}\left[\text{log}\,p(r_t|z,C_t)\right] - \text{KL}(q(z|d_t)||p(z|C_t)) \\
		&\le\text{log}\,\int p(r_t|z,C_t)p(z|C_t)dz \\
		&=\text{log}\,p(r_t|C_t)
	\end{split}
\end{equation}
where $\mathcal{L}$ is called evidence lower bound (ELBO) and $q(z|d_t)$ is called inference network. The learning process of the inference network is shown in Fig.~\ref{fig:fig3}(b).

Similar to the prior network $p(z|C_t)$, the inference network $q(z|d_t)$ approximates the mean and variance of the posterior $p(z|d_t)$ as follows:
\begin{equation}
	\small
	\setlength{\abovedisplayskip}{0.35cm} 
	\setlength{\belowdisplayskip}{0.35cm}
	\left[\begin{array}{c} 
		\mu' \\
		log(\sigma'^2) \\
	\end{array}\right] = \text{MLP}(E(C_t)\oplus E(r_t))
\end{equation}
where $E(C_t)$ and $E(r_t)$ denote the representations of dialogue context and human response in current turn, respectively. We use the reparametrization trick to sample $z$ from the inference network and maximize the ELBO by gradient ascent on a Monte Carlo approximation of the expectation.

It is worth noting that tricks such as mixing $d_t$ with the instances in the data pool and updating IDS for a small number of epochs~\cite{shen2017deep} can be easily adopted to increase the utilization  of labeled data. But, in our experiments, we find there is still a great improvement without these tricks. To reduce  computation load, we update IDS with each $d_t$ only once in a stream-based fashion and leave these tricks in our future work.

\section{Construction of Experimental Data}
To simulate the new unconsidered user needs, one possible method is to delete some question types in the training set of existing datasets~(e.g., bAbI tasks~\cite{bordes2016learning}) and test these questions in the testing phase. However, the dialogue context plays an important role in the response selection. Simply deleting some turns of a dialogue will result in a different system response. For example, in bAbI Task5, deleting those turns on updating api calls will result in a different recommended restaurant. Thus, we do not modify existing datasets but construct a new benchmark dataset to study the inconsistency of training and testing in task-oriented dialogue systems.

We build this dataset based on the following two principles. First of all, we ensure all interactions are reasonable. To achieve that, we follow the construction process of existing work~\cite{bordes2016learning} and generate the dataset by complicated and elaborated rules. Second, the dataset should contain several subsets and the dialogue scenarios covered in each subset are incremental. To simulate the new unconsidered user needs, we train the dialogue system on a smaller subset and test it on a more complicated one.

Specifically, our dataset contains five different subsets within the context of customer services. From SubD1 to SubD5, the user needs become richer in each subset, as described below.

\noindent \textbf{SubD1} includes basic scenarios of the customer services in which users can achieve two primary goals. First, users can look up a product or query some attributes of interested products. For example, they can ask ``Is \$entity\_5\$\footnote{We use special tokens to anonymize all private information in our corpus.} still on sales?'' to ask the discount information of \$entity\_5\$. Second, after finding the desired product, users can consult the system about the purchase process and delivery information.

\noindent \textbf{SubD2} contains all scenarios in SubD1. Besides, users can confirm if a product meets some additional conditions. For example, they can ask ``Does \$entity\_9\$ support Android?'' to verify the operating system requirement.

\noindent \textbf{SubD3} contains all scenarios in SubD2. In addition, users can compare two different items. For example, they can ask ``Is \$entity\_5\$ cheaper than \$entity\_9\$?'' to compare the prices of \$entity\_5\$ and \$entity\_9\$.

\noindent \textbf{SubD4} contains all scenarios in SubD3. And there are more user needs related to the after-sale service. For example, users can consult on how to deal with network failure and system breakdown.

\noindent \textbf{SubD5} contains all scenarios in SubD4. Further more, users can give emotional utterances. For example, if users think our product is very cheap, they may say ``Oh, it's cheap and high-quality. I like it!''. The dialogue system is expected to reply emotionally, such as ``Thank you for your approval.''. If the user utterance contains both emotional and task-oriented factors, the system should consider both. For example, if users say ``I cannot stand the old operating system, what should I do to update it?'', the dialogue system should respond ``I'm so sorry to give you trouble, please refer to this: \$api call update system\$.''.

It is worth noting that it often requires multiple turns of interaction to complete a task. For example, a user wants to compare the prices of \$entity\_5\$ and \$entity\_9\$, but not explicitly gives the two items in a single turn. To complete the missing information, the system should ask which two products the user wants to compare. Besides, the context plays an important role in the dialogue. For example, if users keep asking the same product many times consecutively, they can use the subject ellipsis to query this item in the current turn and the system will not ask users which product they are talking about. In addition, taking into account the diversity of natural language, we design multiple templates to express the same intention. The paraphrase of queries makes our dataset more diverse. For each sub-dataset, there are 20,000 dialogues for training and 5,000 dialogues for testing. A dialogue example in SubD5 and detailed data statistics are provided in the Appendices~\ref{sec:appendix}.

\section{Experimental Setup}
\subsection{Data Preprocessing}
It is possible for the dialogue model to retrieve responses directly without any preprocessing. However, the fact that nearly all utterances contain entity information would lead to a slow model convergence. Thus, we replace all entities with the orders in which they appear in dialogues to normalize utterances. For example, if the \$entity\_9\$ is the second distinct entity which appears in a dialogue, we rename it with \$entity\_order\_2\$ in the current episode. After the preprocessing, the number of normalized response candidates on both the training and test sets in each sub-dataset is shown in Table~\ref{table:table1}.

\begin{table}[!h]
	\small
	\setlength{\abovecaptionskip}{0.1cm}
	\setlength{\belowcaptionskip}{-0.2cm}
	\centering
	\begin{tabular}{@{}l|ccccc@{}}
		\toprule
		          & SubD1 & SubD2 & SubD3 & SubD4 & SubD5 \\ \midrule
		\# of RSP & 41    & 41    & 66    & 72    & 137   \\ \bottomrule
	\end{tabular}
	\caption{The number of normalized response candidates in each sub-dataset after entity replacement, both training and test data included.}
	\label{table:table1}
\end{table}

\subsection{Baselines}
We compare IDS with several baselines:
\begin{itemize}
	\item {IR}: the basic tf-idf match model used in~\cite{bordes2016learning,dodge2015evaluating}.
	\item {Supervised Embedding Model (SEM)}: the supervised word embedding model used in~\cite{bordes2016learning,dodge2015evaluating}.
	\item {Dual LSTM (DLSTM)}: the retrieval-based dialogue model used in~\cite{lowe2015ubuntu}.
	\item {Memory Networks (MemN2N)}: the scoring model which is used in QA~\cite{sukhbaatar2015end} and dialogue systems~\cite{bordes2016learning,dodge2015evaluating}.
	\item {IDS$^-$}: IDS without updating model parameters during testing. That is, IDS$^-$ is trained only with human intervention data on the training set and then we freeze parameters.
\end{itemize}

\subsection{Measurements}
Following the work of \citeauthor{williams2017hybrid}~\shortcite{williams2017hybrid} and \citeauthor{bordes2016learning}~\shortcite{bordes2016learning}, we report the average turn accuracy. The turn is correct if the dialogue model can select the correct response, and incorrect if not. Because IDS requires human intervention to reduce risks whenever there is low confidence, we calculate the average turn accuracy \emph{only if} IDS chooses to respond without human intervention. That is, compared with baselines, IDS computes the turn accuracy only on a subset of test sets. To be fair, we also report the rate at which IDS refuses to respond on the test set. The less the rejection rate is, the better the model performs.

\subsection{Implementation Details}
Our word embeddings are randomly initialized. The dimensions of word embeddings and GRU hidden units are both 32. The size of the latent variable $z$ is 20. In uncertainty estimation, the  repetition time $K$ is 50. In all experiments, the average $\text{JSD}$ threshold $\tau_1$ and the response probability threshold $\tau_2$ are both set to 0.3\footnote{The smaller $\tau_1$ or larger $\tau_2$ will result in a higher average turn accuracy but a larger human intervention frequency. In our preliminary experiments, we find that setting both $\tau_1$ and $\tau_2$ to 0.3 is a good trade-off.}. In online learning, the number of Monte Carlo sampling is 50. In all experiments, we use the ADAM optimizer~\cite{kingma2014adam} and the learning rate is 0.001. We train all models in mini-batches of size 32.

\section{Experimental Results}
\subsection{Robustness to Unconsidered User Actions}
To simulate unexpected user behaviors after deployment, we use the hardest test set, SubD5, as the common test set, but train all models on a simple dataset~(SubD1-SubD4) individually. The average turn accuracy is shown in Table~\ref{table:table2}.

\begin{table}[!t]
	\small
	\centering
	\setlength{\abovecaptionskip}{0.1cm}
	\setlength{\belowcaptionskip}{-0.1cm}
	\begin{tabular}{@{}l|cccc@{}}
		\toprule
		& \multicolumn{4}{c}{\textbf{Training DataSet}}                            \\
		\textbf{Model} & SubD1           & SubD2           & SubD3           & SubD4           \\ \midrule
		IR             & 34.7\%          & 35.2\%          & 44.0\%          & 55.1\%          \\
		SEM            & 35.1\%          & 35.4\%          & 43.4\%          & 52.7\%          \\
		DLSTM          & 48.2\%          & 52.0\%          & 61.7\%          & 74.0\%          \\
		MemN2N         & 50.5\%          & 50.4\%          & 64.0\%          & 77.4\%          \\ \midrule
		IDS$^-$         & 78.6\%          & 77.3\%          & 83.2\%          & 92.7\%          \\
		IDS          & \textbf{98.1\%} & \textbf{96.7\%} & \textbf{99.0\%} & \textbf{99.7\%} \\ \bottomrule
	\end{tabular}
	\caption{The average turn accuracy of different models. Models are trained on SubD1-SubD4 respectively, but all tested on SubD5. Note that, unlike the existing methods, IDS$^-$ and IDS give responses only if there is high degree of confidence.}
	\label{table:table2}
\end{table}
\begin{table}[!t]
	\small
	\centering
	\setlength{\abovecaptionskip}{0.1cm}
	\setlength{\belowcaptionskip}{-0.1cm}
	\begin{tabular}{@{}l|cccc@{}}
		\toprule
		& \multicolumn{4}{c}{\textbf{Training DataSet}} \\
		\textbf{Model} & SubD1     & SubD2    & SubD3    & SubD4       \\ \midrule
		IDS$^-$         & 42.0\%    & 35.5\%   & 30.4\%   & 32.0\%   \\
		IDS          & 79.4\%    & 79.0\%   & 66.6\%   & 62.8\%   \\ \bottomrule
	\end{tabular}
	\caption{The rejection rate on the test set of SubD5.}
	\label{table:table3}
\end{table}

When trained on SubD1 to SubD4 and tested on SubD5, as shown in Table~\ref{table:table2}, the existing methods are prone to poor performance because these models are not aware of which instances they can handle. However, equipped with the uncertainty estimation module, IDS$^-$ can refuse to respond the uncertain instances and hence achieves better performance. For example, when trained on SubD1 and tested on SubD5, IDS$^-$ achieves 78.6\% turn accuracy while baselines achieve only 50.5\% turn accuracy at most. Moreover, if updating the model with human intervention data during testing, IDS attains nearly perfect accuracy in all settings.

Due to the uncertainty estimation module, IDS$^-$ and IDS will refuse to respond if there is low confidence. The rejection rates of them are shown in Table~\ref{table:table3}.
The rejection rate will drop if  the training set is similar to the test set. Unfortunately, the rejection rate of IDS is much higher than that of IDS$^-$. We guess the reason is the catastrophic forgetting~\cite{french1999catastrophic,kirkpatrick2017overcoming}. When IDS learns to handle new user needs in SubD5, the knowledge learnt in the training phase will be somewhat lost. Thus, IDS needs more human intervention to re-learn the forgotten knowledge. However, forgetting will not occur if IDS is deployed from scratch and accumulates knowledge online because weights of IDS are optimized alternatively on all possible user needs.

\subsection{Deploying without Initialization}
\label{efficiency}
Compared with existing methods, IDS can accumulate knowledge online from scratch. The uncertainty estimation module will guide us to label only valuable data. This is similar to {\em active learning}~\cite{balcan2009agnostic,dasgupta2005analysis}.

To prove that, we train baselines on each of the SubD$i$ training data with one epoch of back propagation\footnote{In the online learning process of IDS$^-$, each labeled data in the data pool is used only once. For the sake of fairness, we train baselines with only one epoch in this section.} and test these models on each of the SubD$i$ test set. In contrast, for each SubD$i$ training set, IDS$^-$ is trained from \emph{random initialization}. Whenever IDS$^-$ refuses to respond, the current context-response pair in the training set will be used to update the model until all training data in SubD$i$ are finished. Hence IDS$^-$ is trained on the subset of SubD$i$ where the response confidence is below the threshold. After the training is finished, we freeze the model parameters and test IDS$^-$ on the test set of SubD$i$.

\begin{table}[!t]
	\centering
	\small
	\setlength{\abovecaptionskip}{0.1cm}
	\setlength{\belowcaptionskip}{-0.2cm}
	\setlength{\tabcolsep}{2mm}
	{
		\begin{tabular}{@{}l|ccccc@{}}
			\toprule
			\textbf{Model} & SubD1          & SubD2          & SubD3          & SubD4           & SubD5           \\ \midrule
			IR             & 66.3\%         & 66.5\%         & 70.8\%         & 74.1\%          & 75.7\%          \\
			SEM            & 67.6\%         & 68.4\%         & 64.1\%         & 60.8\%          & 65.8\%          \\
			DLSTM          & 99.9\%         & 99.9\%         & 98.8\%         & 97.7\%          & 96.7\%          \\
			MemN2N         & 93.4\%         & 94.5\%         & 89.8\%         & 85.3\%          & 80.8\%          \\ \midrule
			IDS$^-$          & \textbf{100\%} & \textbf{100\%} & \textbf{100\%} & \textbf{99.8\%} & \textbf{99.9\%} \\ \bottomrule
		\end{tabular}
	}
	\caption{The average turn accuracy of different systems on SubD$i$ test set. Note each baseline is trained on the entire SubD$i$ training data, but IDS$^-$ is trained only on the low-confidence subset of SubD$i$ training set. The parameters of all system are frozen during testing.}
	\label{table:table4}
\end{table}

\begin{table}[!t]
	\centering
	\small
	\setlength{\abovecaptionskip}{0.1cm}
	\setlength{\belowcaptionskip}{-0.2cm}
	\begin{tabular}{@{}ccccc@{}}
		\toprule
		SubD1  & SubD2  & SubD3  & SubD4  & SubD5  \\ \midrule
		24.1\% & 27.4\% & 38.4\% & 56.5\% & 61.6\% \\ \bottomrule
	\end{tabular}
	\caption{The rejection rate of IDS$^-$ on SubD$i$ training set.}
	\label{table:table5}
\end{table}

\begin{table}[!t]
	\small
	\setlength{\abovecaptionskip}{0.1cm}
	\setlength{\belowcaptionskip}{-0.2cm}
	\centering
	\begin{tabular}{@{}ccccc@{}}
		\toprule
		SubD1     & SubD2    & SubD3      & SubD4      & SubD5    \\ \midrule
		0.3\%     & 0.7\%    & 3.2\%      & 13.8\%     & 24.1\%   \\ \bottomrule
	\end{tabular}
	\caption{The rejection rate of IDS$^-$ on SubD$i$ test set.}
	\label{table:table6}
\end{table}

Table~\ref{table:table4} shows the average turn accuracy of different models. Table~\ref{table:table5} shows the rejection rate of IDS$^-$ on each SubD$i$ training set. We see that, compared with all baselines, IDS$^-$ achieves better performance with much less training data. This shows the uncertainty estimation module can select the most valuable data to label online.

Table~\ref{table:table6} shows the rejection rate of IDS$^-$ on each SubD$i$ test data. We can see that the rejection rate is negligible on SubD1, SubD2 and SubD3. It means IDS$^-$ can converge to a low rejection rate after deployment. For SubD4 and SubD5, there are still some instances IDS$^-$ can not handle.  It is due to the fact that SubD4 and SubD5 are much more complicated than others. In the next section, we further show that as online learning continues, the rejection rate will continue to drop as well.

\subsection{Frequency of Human Intervention}
The main difference between our approach and others is that we introduce humans in the system loop. Therefore, we are interested in the question of how frequently humans intervene over time. 

\begin{figure}[!t]
	\centering
	\setlength{\abovecaptionskip}{0.1cm}
	\setlength{\belowcaptionskip}{-0.2cm}
	\includegraphics[width=0.43\textwidth]{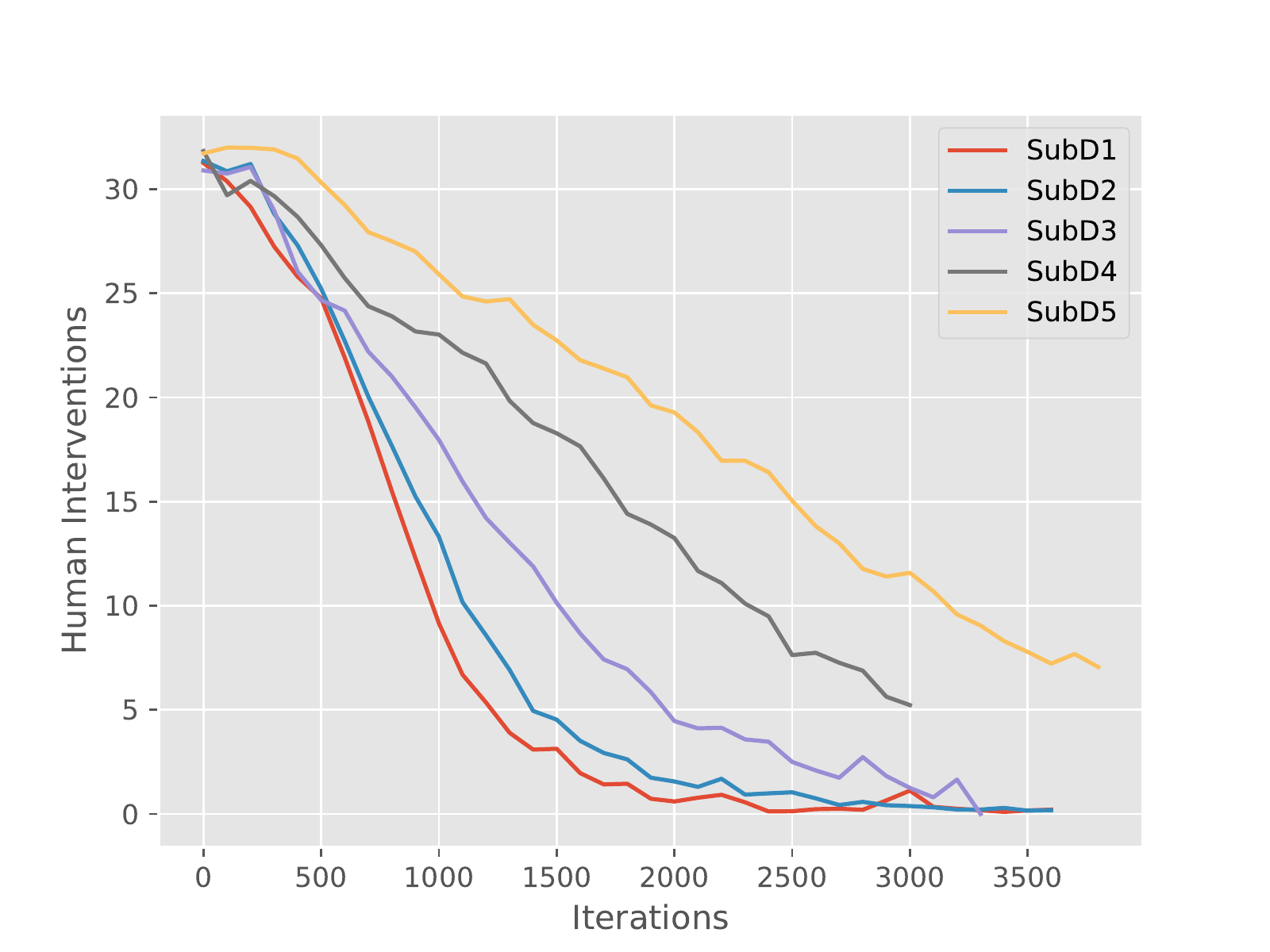}
	\caption{The intervention frequency curves after deploying IDS$^-$ without any initialization.}
	\label{fig:fig5}
\end{figure}

The human intervention frequency curves of deploying IDS$^-$ without any initialization~(i.e., the online learning stage of IDS$^-$ in Section~\ref{efficiency}) are shown in Fig.~\ref{fig:fig5}. As shown, the frequency of human intervention in a batch will decrease with time. In the early stage of deployment, IDS$^-$ has a large degree of uncertainty because there are only a few context-response pairs in the data pool. Through continuous interactions with users, the labeled data covered in the data pool will become more and more abundant. Thus, humans are not required to intervene frequently.

Besides, human intervention curves of different datasets have different convergence rates. The curve of SubD1 has the fastest convergence rate. As the dataset covers more and more user needs, the convergence rate becomes slower. However, there is still a trend to converge for SubD4 and SubD5 as long as we continue the online learning. This phenomenon is in line with the intuition that a more complicated dialogue system requires more training data than a simple one.

\begin{figure}[!t]
	\centering
	\setlength{\abovecaptionskip}{0.1cm}
	\setlength{\belowcaptionskip}{-0.2cm}
	\subfigure{
		\begin{minipage}[b]{0.46\linewidth }
			\includegraphics[width=1\linewidth]{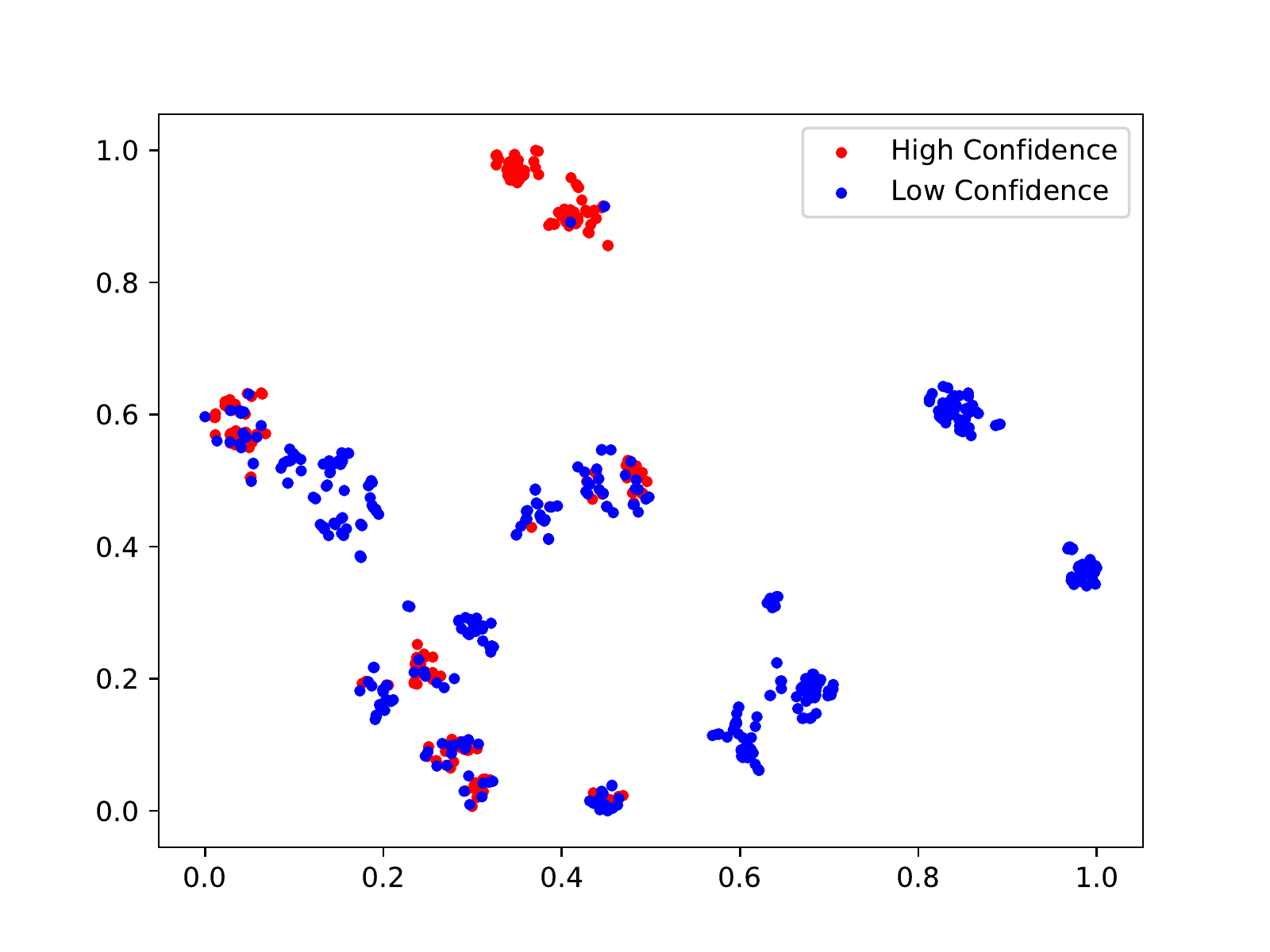}
			\includegraphics[width=1\linewidth]{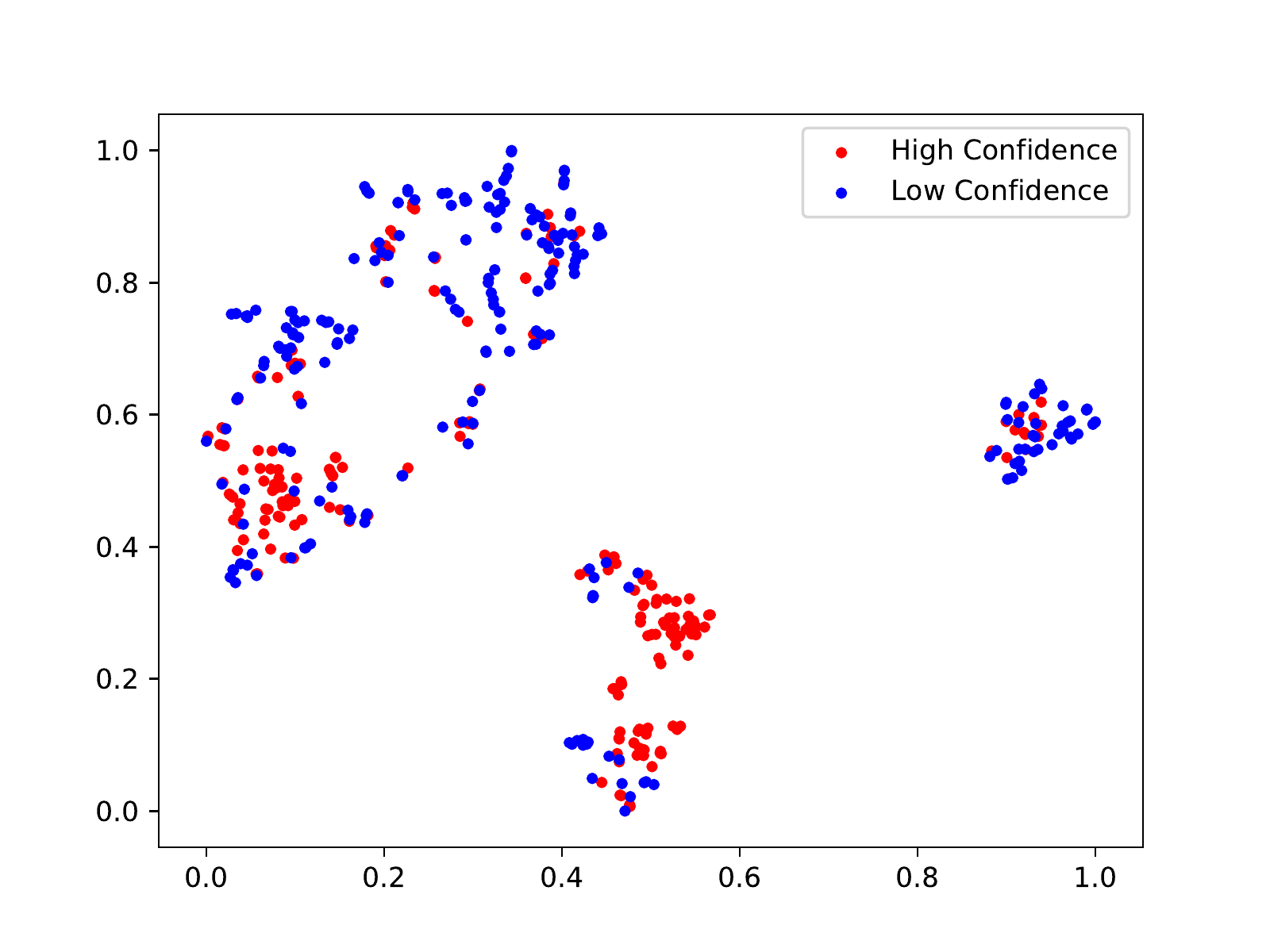}
		\end{minipage}
	}
	\subfigure{
		\begin{minipage}[b]{0.46\linewidth }
			\includegraphics[width=1\linewidth]{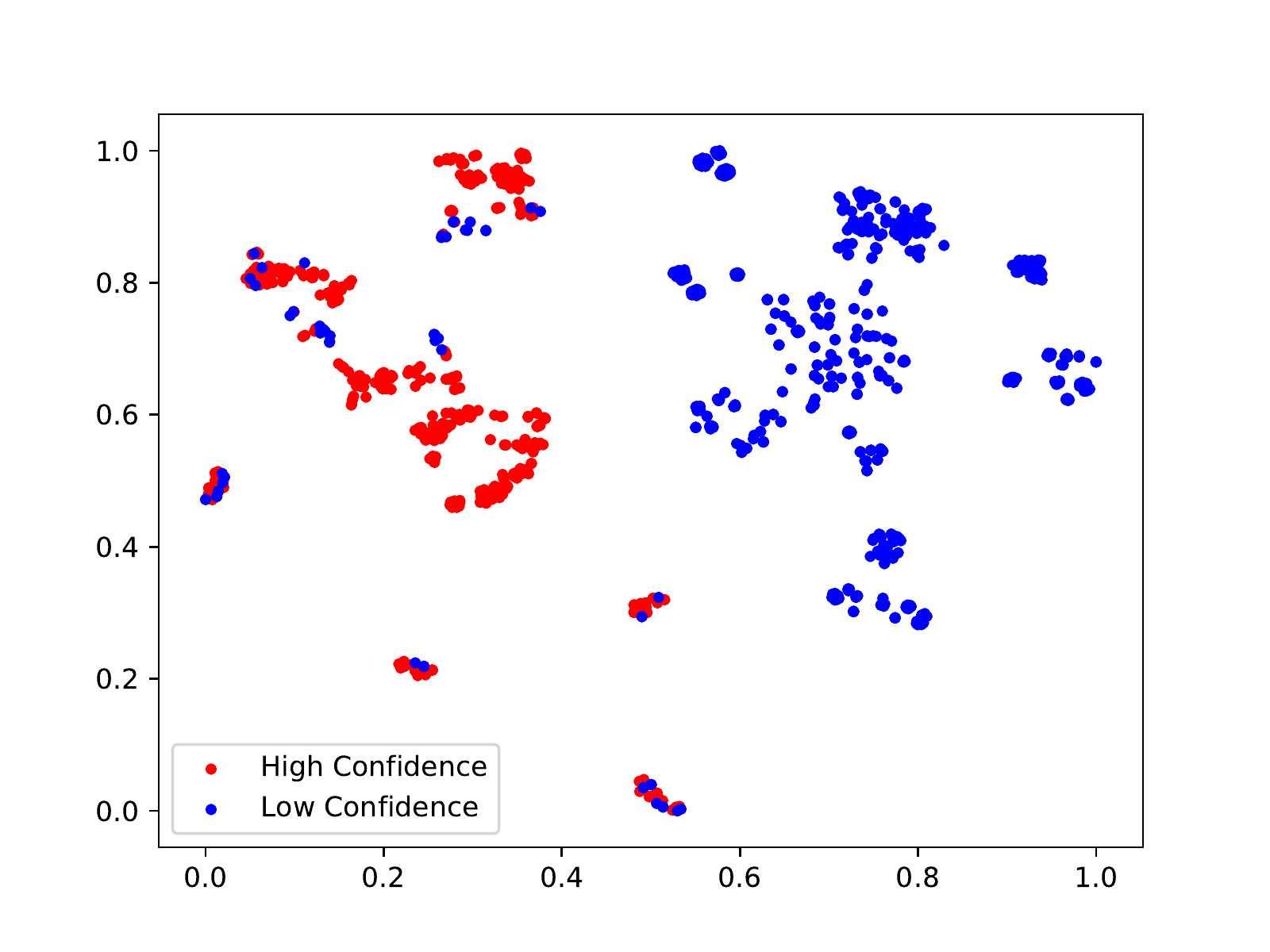}
			\includegraphics[width=1\linewidth]{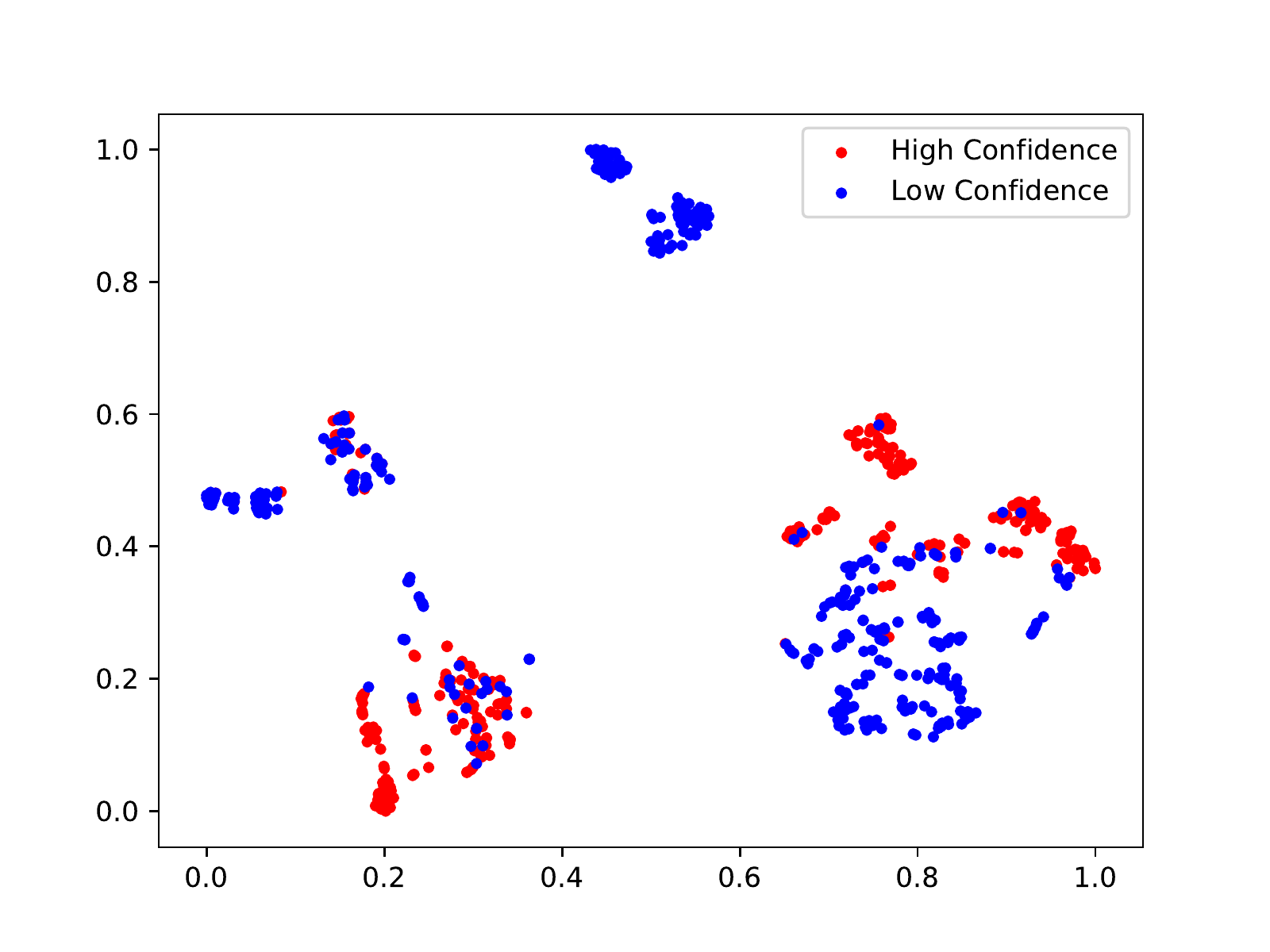}
		\end{minipage}
	}
	\caption{t-SNE visualization on the context representations of four different system responses. Red dots are contexts responded by IDS$^-$ with high confidence, while blue dots are contexts with low confidence.}
	\label{fig:fig6}
\end{figure}

\subsection{Visual Analysis of Context Embedding}
To better understand the behavior of our approach, we train IDS$^-$ on the SubD5 training set until 2,000 batches online updates are finished, and then freeze the model parameters and test it on the SubD5 test set. As Table~\ref{table:table1} shows, there are 137 unique normalized responses. Among these responses, we pick four of them and draw their context embedding vectors. Each vector is reduced to a 2-dimensional vector via t-SNE~\cite{maaten2008visualizing} for visualization, one sub-graph per response in Fig.~\ref{fig:fig6}. In each figure, the red dots are contexts responded by IDS$^-$ with high confidence, while the blue dots are contexts responded by human where there is low confidence.

These graphs show a clear separation of sure vs. unsure contexts. Some blue dots are far away from the red. Humans should pay attention to these contexts to avoid risks. Besides, there are only a small number of cases when the two classes are mingled. We guess these cases are located in the confidence boundary. In addition, there are multiple clusters in each class. It is due to the fact the same system response can appear in different dialogue scenes. For example, ``the system requesting user's phone number'' appears in scenes of both exchange and return goods. Although these contexts have the same response, their representations should be different if they belong to different dialogue scenes.

\section{Related Work}
Task-oriented dialogue systems have attracted numerous research efforts. Data-driven methods, such as reinforcement learning~\cite{williams2017hybrid,zhao2016towards,li2017end} and supervised learning~\cite{wen2016network,Eric2017Key,bordes2016learning}, have been applied to optimize dialogue systems automatically. These advances in task-oriented dialogue systems have resulted in impressive gains in performance. However, prior work has mainly focused on building task-oriented dialogue systems in a closed environment. Due to the biased assumptions of real users, such systems will break down when encountering unconsidered situations.

Several approaches have been adopted to address this problem. \citeauthor{gavsic2014incremental}~\shortcite{gavsic2014incremental} explicitly defined kernel functions between belief states from different domains to extend the domain of dialogue systems. But it is difficult to define an appropriate kernel function when the ontology has changed drastically. \citeauthor{shah2016interactive}~\shortcite{shah2016interactive} proposed to integrate turn-level and task-level reward signals to learn how to handle new user intents. \citeauthor{lipton2018bbq}~\shortcite{lipton2018bbq} proposed to use BBQ-Networks to extend the domain. However, \citeauthor{shah2016interactive}~\shortcite{shah2016interactive} and \citeauthor{lipton2018bbq}~\shortcite{lipton2018bbq} have reserved a few bits in the dialogue state for the domain extension. To relax this assumption, \citeauthor{wang2018teacher}~\shortcite{wang2018teacher} proposed the teacher-student framework to maintain dialogue systems. In their work, the dialogue system can only be extended offline after finding errors and it requires hand-crafted rules to handle new user actions. In contrast, we can extend the system online in an incremental\footnote{The term ``incremental'' refers to systems able to operate on a word by word basis in the previous work~\cite{eshghi2017bootstrapping,schlangen2009general}. In our work, it refers to the system which can adapt to new dialogue scenarios after deployment.} way with the help of hired customer service staffs.

Our proposed method is inspired by the cumulative learning~\cite{fei2016learning}, which is a form of lifelong machine learning~\cite{chen2016lifelong}. This learning paradigm aims to build a system that learns cumulatively. The major challenges of the cumulative learning are finding unseen classes in the test set and updating itself efficiently to accommodate new concepts~\cite{fei2016learning}. To find new concepts, the heuristic uncertainty estimation methods~\cite{tong2001support,culotta2005reducing} in active learning~\cite{balcan2009agnostic,dasgupta2005analysis} can be adopted. When learning new concepts, the cumulative learning system should avoid retraining the whole system and catastrophic forgetting~\cite{french1999catastrophic,kirkpatrick2017overcoming}. But the catastrophic forgetting does not happen if the dialogue system is trained with all possible user needs alternatively from scratch.

The uncertainty estimation and online learning methods in our work are inspired by variational inference approach~\cite{Rezende2014Stochastic,Kingma2014Auto}. In the existing work, this approach was used to generate diverse machine responses in both open domain dialogue systems~\cite{zhao2017learning,Serban2016A} and task-oriented dialogue systems~\cite{wen2017latent}. In contrast, our work makes use of the Bayesian nature of variational inference to estimate the uncertainty and learn from humans. Specifically, we sample variables from the prior network as the random perturbation to estimate the model uncertainty following the idea of Query-By-Committee~\cite{seung1992query} and optimize model parameters by maximizing the ELBO.

\section{Conclusion}
This paper presents a novel incremental learning framework to design dialogue systems, which we call IDS. In this paradigm, users are not expected to follow any definition, and IDS has potential to handle new situations. To simulate new user actions after deployment, we propose a new dataset consisting of five different subsets. Experiments show that IDS is robust to new user actions. Importantly, with humans in the loop, IDS requires no data for initialization and can update itself online by selecting the most valuable data. As the usage grows, IDS will cumulate more and more knowledge over time.

\section{Acknowledgments}
The research work described in this paper has been supported by the National Key Research and Development Program of China under Grant No. 2017YFB1002103 and the Natural Science Foundation of China under Grant No. U1836221.

\bibliography{acl2019}
\bibliographystyle{acl_natbib}

\appendix

\section{Appendices}
\label{sec:appendix}
\subsection{Dialogue Example}
Our corpus is in Chinese. We translate an example dialogue into English and show it in Table~\ref{table:table7}.

\begin{table*}[!t]
	\centering
	\begin{tabular}{@{}ll@{}}
		\toprule
		\textit{User:}                    & The operating system breaks down. What should I do?              \\ \midrule
		\textit{System:}                  & You can try to update the system.                                \\ \midrule
		\textit{User:}                    & The system always has problems. I'm very disappointed.           \\ \midrule
		\textit{System:}                  & I'm so sorry to give you trouble, we will do our best to improve the operating system. \\ \midrule
		\textit{User:}                    & I don't know how to update the system.                           \\ \midrule
		\textit{System:}                  & Please refer to this: \$api call update system\$.                \\ \midrule
		\textit{User:}                    & It looks so troublesome. I want to send the product back.           \\ \midrule
		\textit{System:}                  & Please tell me your order number.                                \\ \midrule
		\textit{User:}                    & My order number is \$orderNumber\$.                              \\ \midrule
		\textit{System:}                  & Please send the item to this address: \$address\$.               \\ \bottomrule
	\end{tabular}
	\caption{A sample dialogue in SubD5. The dialogue consists of two dialogue scenarios: ``consult system error'' and ``return goods''. And in the second turn, the user expressed negative emotion. Besides, it is worth noting that different user needs involve different business logics. For example, if the user wants to return goods, the system need know the order number and ask the user to send the goods to the \$address\$.}
	\label{table:table7}
\end{table*}

\subsection{Data Statistics}
Data statistics are show in Table~\ref{table:table8}. Dialogue scenarios in each sub-dataset are shown in \ref{table:table9}.
\begin{table*}[!h]
	\centering
	\setlength{\tabcolsep}{3mm}
	{
		\begin{tabular}{@{}l|ccccc@{}}
			\toprule
			& SubD1 & SubD2 & SubD3 & SubD4 & SubD5 \\ \midrule
			\# of Scenarios per Dialogue    & 5.2         & 5.2         & 4.6         & 3.7         & 3.9         \\ \midrule
			\# of Utterances per Dialogue   & 11.7        & 11.7        & 10.6        & 9.8         & 12.4        \\ \midrule
			\# of Tokens per Utterance      & 3.8         & 4.0         & 4.1         & 4.3         & 5.1         \\ \midrule
			\# of Paraphrases per Query     & 8.9         & 7.0         & 6.5         & 6.9         & 6.9         \\ \midrule
			Vocab Size after Preprocessing  & 194         & 253         & 303         & 430         & 620         \\ \midrule
			\# of Products                  & \multicolumn{5}{c}{50}                                              \\ \midrule
			Training Dialogues              & \multicolumn{5}{c}{20000}                                           \\ \midrule
			Validation Dialogues            & \multicolumn{5}{c}{5000}                                            \\ \midrule
			Test Dialogues                  & \multicolumn{5}{c}{5000}                                            \\ \bottomrule
		\end{tabular}
	}
	\caption{Data statistics of each sub-dataset.}
	\label{table:table8}
\end{table*}

\begin{table*}[!h]
	\centering
	\begin{tabular}{@{}l|l@{}}
		\toprule
		SubD1                  & query product information, query payment methods, query express information   \\\midrule
		SubD2                  & \textit{scenarios of SubD1}, verify product information \\ \midrule
		SubD3                  & \textit{scenarios of SubD2}, compare two products       \\ \midrule
		\multirow{2}{*}{SubD4} & \textit{scenarios of SubD3}, ask for an invoice, consult system error, consult nfc error, \\
		                       & consult network error, return goods, exchange goods, query logistics                      \\ \midrule
		SubD5                  & \textit{scenarios of SubD4}, express positive emotion, express negative emotion           \\ \bottomrule
	\end{tabular}
	\caption{The dialogue scenarios covered in each sub-dataset.}
	\label{table:table9}
\end{table*}

\end{document}